# Machine Biometrics - Towards Identifying Machines in a Smart City Environment

George K. Sidiropoulos, George A. Papakostas
HUman-MAchines INteraction Laboratory (HUMAIN-Lab), Department of Computer Science, International Hellenic University
Kavala, Greece
{georsidi, gpapak}@cs.ihu.gr

*Abstract*—This paper deals with the identification of machines in a smart city environment. The concept of *machine biometrics* is proposed in this work for the first time, as a way to authenticate machine identities interacting with humans in everyday life. This definition is imposed in modern years where autonomous vehicles, social robots, etc. are considered active members of contemporary societies. In this context, the case of car identification from the engine behavioral biometrics is examined. For this purpose, 22 sound features were extracted and their discrimination capabilities were tested in combination with 9 different machine learning classifiers, towards identifying 5 car manufacturers. The experimental results revealed the ability of the proposed biometrics to identify cars with high accuracy up to 98% for the case of the Multilayer Perceptron (MLP) neural network model.

*Keywords—biometrics, machine learning, artificial intelligence, smart city, identity authentication*

## I. Introduction

With the advent of technology and the exponential increase in the application of human-computer interfaces due to the Industry 4.0, machines and robots have already become part of the everyday life of humans [1]. With this growing usage of technology, identity authentication has become a top priority in many cases, with a large number of studies proposing new methodologies or new biometrics for authentication. Even before those advancements, the authentication of a person was applied in many scenarios, like facilities or licenses.

As the development of Industry 4.0 continues, so has the shift and development towards smart cities [2]. A smart city aims to improve the quality of life in the urban areas by integrating information techniques (big data, cloud computing, Internet of Things, etc.) and build various smart environments [3]. As a result, unauthorized access to those networks and environments is a critical issue that has been addressed through various authentication processes. Therefore, identity authentication, as a security service, provides the user access to those services.

Moreover, security and rapid advancements in networking and communication have increased the need for reliable identity authentication [4]. The three types of authentication elements (password, hardware equipment and biometrics) address these security issues. For this reason, many security measures have been proposed, with the discipline of biometrics, which focuses on establishing the identity of an individual based on the inherent physical or behavioral traits [5], [6].

As mentioned before, smart cities are composed of many small smart environments, such as smart homes, smart libraries, intelligent transportation systems or smart healthcare [7]. The ability to access those environments at any time is an important foundation for a smart city, which, as a result, unauthenticated access can pose serious problems to those kinds of network pipelines. Therefore, secure authentication processes are required and are mostly tackled by the usage of biometric features. Biometric features include physiological features, such as the face, fingerprints, or iris, and behavioral features, such as gait, voice, or signature. These can be used to discriminate one individual from another [8] for access control or in forensics and unlike other types of authentication processes, they cannot be stolen, lost or forgotten.

The contribution of this paper is twofold. Firstly, a new category of biometrics, named *machine biometrics* is proposed for the first time, for machines identification. As for humans, machine biometrics are unique measurements related to a machine's characteristics, that aim to identify a particular machine or type of machine. Secondly, the *engine biometrics* based on the sound of a car's engine is proposed for the identification of passenger vehicles (cars).

The remainder of the paper is organized as follows. Section II introduces the new paradigm of *machine biometrics*; Section III describes the proposed *engine biometric*, Section IV presents the experimental study on the performance of the *engine biometrics*. Finally, Section V concludes this study and points out the future work.

## II. Machine Biometrics – A New Paradigm

The previous section already mentioned the emerging trends coming from the Industry 4.0 framework, where cyber-physical systems are interacting with each other and with humans. In this context, smart cities are considered the future human settlements integrating the recent technological innovations. In such environments, autonomous vehicles on a city's streets, social robots accompanying people in urban districts will constitute typical examples of highly intelligent machines integration in human societies.

The increased presence of autonomous and intelligent machines in daily life activities, calls for the establishment of a framework for their monitoring, supervision and authentication. Therefore, it is the right time to define the biometrics of machines in full proportion to the biometrics of humans.

**Definition 1** – A *machine biometric* is a set of measurements that describe the inherent characteristics of a machine.

Following the same methodology with human's biometric characteristics, seven factors [5] that define the suitability of a biometric trait in order to be useful for identity authentication can be adopted: (1) **Universality**, which in the case of *machine biometrics* can be defined per machine type, (2) **Uniqueness**, to distinguish one machine from another, (3) **Permanence**, to remain unchanged over time, (4) **Measurability**, to be acquired by the sensor networks that are part of a smart city, (5) **Performance**, to be accurate and sustainable in terms of the used resources, (6) **Acceptability**, where the machine enables the measurement of its biometric, without any permission and (7) **Circumvention**, to avoid any unauthorized machine imitating a different identity.

In order to describe a *machine biometric* based on the previously presented theory, the concept of *machine identity* should be defined firstly. Although there are several definitions of human's identity, from different disciplines (sociology, psychology, neuroscience, etc.), for the case of the machines, the following definition can be stated:

**Definition 2** – *Machine identity* refers to the set of qualities and characteristics that the machine has from the construction or formed during its operation and are able to differentiate it in relation to its peers.

Based on the above definitions several *machine biometrics* can be defined depending on the machine type. An example of a *machine biometric* is proposed in the next section for the case of passenger vehicles (cars).

### III. PROPOSED MACHINE BIOMETRIC

In our study, we propose the sound a car's engine generates, as an e*ngine biometric*, which is used with Machine Learning (ML) for identifying a car. More precisely, a fitting application of the proposed *machine biometric* characteristic is the recognition of a car's manufacturer by using the sound the engine generates during its operation. Figure 1, shows the pipeline of the proposed identification methodology with the main function of each processing step being as follows:

1. **Data Gathering:** The sounds of different car engines from different manufacturers were recorded.
2. **Feature Extraction:** This step includes the extraction of the required features from the recordings, by processing the waveform in segments.
3. **Car manufacturer prediction:** In this step, the car's manufacturer is predicted.

*A. Data Gathering*

The data used in this work were gathered by recording the sound of the running engine of cars from different manufacturers. Each car engine was recorded in three different rounds per minute (rpm) states, specifically at: 1000 (idle), 1500 and 2000 rpm.

*B. Feature Extraction*

As depicted in Fig. 1, the waveform of each recording was split into segments using a non-overlapping sliding window. The size of the window was determined for each car engine recording separately, by firstly calculating its tempo and its duration in samples (Eq. 1) as follows:

$$Samples\ per\ tempo = 60/tempo * sample\ rate \quad (1)$$

where the tempo is calculated by using the onset strength of the waveform and the sample rate is a constant value of 44100 (the sample rate of all the recordings).

A total of 22 features were extracted for each waveform window, namely: RMS [9], Zero Crossing [10], Chroma Cens [11], Chroma Short-Time Fourier Transform (STFT) [11], Chroma CQT [12], Spectral Centroid [13], Spectral Bandwidth [14], Spectral Contrast [15], Spectral Flatness [16], Spectral Roll-off [17], Polynomial Zero, Polynomial Linear, Polynomial Quadratic, Mel Frequency Cepstral Coefficient (MFCC) [18], Mel Spectrogram, Spectral Flux [19], Superflux [20], Tonnetz [21], Tempogram [22], Filterbank Energies [23], Log Filterbank Energies and Spectral Subband Centroids [24].

It should be noted that the mean values for each feature were used for the simulations in the study. Additionally, the features were extracted by using the Librosa [25] and Python Speech Features [26] libraries.

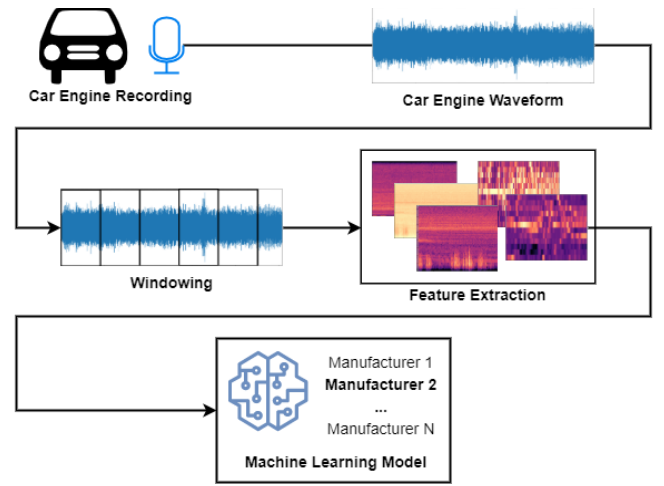

Fig. 1. Pipeline of the proposed methodology.

*C. Car manufacturer prediction*

The features from the previous step were used to train the following 9 ML models:

1. **Linear Support Vector Classification (LSVC):** A similar implementation to the famous Support Vector Machine with a linear kernel, used for classification [27].
2. **Decision Tree (DT):** A predictive modelling approach that uses a decision tree to go from observations to conclusions [28].
3. **K-Nearest Neighbors (KNN):** A non-parametric classification method, with the input consisting of the *k* closest training examples and the output, is based on its neighbors [29].
4. **Logistic Regressor (LR):** A statistical model that uses a logistic function to model dependent variables [30].
5. **Multilayer Perceptron (MLP):** A feedforward Artificial Neural Network [31].
6. **Random Forest (RF):** An ensemble learning method that uses a number of decision tree classifier [32].

7. **Stochastic Gradient Descent Classifier (SGD):** Iterative method for optimizing an objective function [33].
8. **XGBoost Classifier (XGB):** An efficient implementation of gradient boosting [34].
9. **XGBoost Random Forest Classifier (XGBRF):** An efficient application of the XGBoost algorithm on the Random Forest classifier.

IV. EXPERIMENTS

A. Dataset and Configuration

For the experiments, a total of 122 recordings were gathered from the engines of 19 different car models, belonging to 5 different car manufacturers, as depicted in Table I.

TABLE I. DATASET INFORMATION

| Car Manufacturers | | | | |
|---|---|---|---|---|
| *Citroen* | *Fiat* | *Ford* | *Opel* | *Peugeot* |
| C3 (2) C4 (2) Saxo (2) | Panda (4) Punto(3) | Fiesta (2) Focus (3) | Astra (6) Corsa (4) | 206 (2) 307 (3) |

As mentioned before, the engines were recorded for three different rpm levels (1000, 1500 and 2000 rpm). The reasoning behind the recording of the engine in different rpm levels is to study if the running speed makes the engine biometric more discriminative.

The sounds of the engines were recorded using an XXL Inside microphone from the XD02 Drum Kit Pack Micro Microphone Package, equipped with a wind noise reduction sponge foam, to reduce environmental noise as much as possible. The microphone was connected to a smartphone via an XLR Microphone Audio Adapter from SmartRig and the Easy Voice Recorder App was used, with the following option settings: High sound quality, ".wav" encoding, 44.1KHz sample rate with noise and echo cancellation being enabled.

To record a sound coming from an engine, the car was left on a neutral gear in all cases and for the cases of 1500 and 2000 rpm, the gas pedal was being pressed so that the engine was running steadily at a specific speed. The engines were recorded with the car's hood being open and by positioning the microphone at the center of the engine, for a duration of 15 seconds.

B. Simulations

In our study, we evaluated the application of 9 ML models for the recognition of the manufacturer of different car engines. The experiments were conducted using the Scikit-learn [35] Library for Python.

*1) Data preparation*

For each sound waveform, a non-overlapping sliding window technique was applied to split the waves into multiple segments. In this study, the performance of the models on different window sizes was examined, specifically with a duration of 1, 2 and 5 tempos, on different engine rpm levels. Thus, the performance of the models was tested on 9 different datasets, with different sizes each due to the varying length of window size. It should be noted that before the training process, the dataset's values were being normalized, so as to preserve their values between the range of [0, 1], which assists with the training process and models generally perform better.

*2) Performance Evaluation*

The models were evaluated using the Accuracy, Precision, Recall and F1-score measures. These measures are widely used in machine learning to evaluate the performance of a classification model.

*3) Hyperparameter optimization*

In order to optimize the performance of each ML model, a hyperparameter optimization technique was employed using the Scikit-Optimize Library [36]. The technique that was employed was the Bayesian optimization scheme over the hyperparameters of each model, which does not try out all the possible parameter values. A fixed number of parameter settings were sampled from specific value distributions, measuring their performance over a 10-fold cross validation technique and selecting the best according to the highest F1-score.

*4) Model performance validation*

After the hyperparameter optimization step was finished, each model's performance was evaluated by applying the Leave-One-Out evaluation strategy for each sample of the dataset.

*5) Results*

The experimental results are depicted in Table II, showing the validation results of the models in the dataset with the car engines running at 2000 rpm with the 1 tempo window size.

TABLE II. BEST PERFORMING VALIDATION RESULTS

| ML Model | Accuracy (%) | Precision (%) | Recall (%) | F1-score (%) |
|---|---|---|---|---|
| LSVC | 74.86 | 74.78 | 76.58 | 75.20 |
| DT | 94.16 | 94.12 | 94.17 | 94.13 |
| KNN | 98.18 | 98.16 | 98.13 | 98.14 |
| LR | 71.86 | 72.22 | 73.12 | 72.46 |
| MLP | **98.50** | **98.45** | **98.44** | **98.45** |
| RF | 98.42 | 98.38 | 98.40 | 98.39 |
| SGD | 72.34 | 72.56 | 73.71 | 73.05 |
| XGB | 98.03 | 97.93 | 98.02 | 97.98 |
| XGBRF | 97.47 | 97.41 | 97.41 | 97.41 |

The experimental results from the validation phase generally show very satisfactory performance from the included models. The MLP model has performed the best, with 98.45% F1-score, followed by the RF and KNN models with 98.39% and 98.14% F1-scores respectively. The LR model seems to have performed the worst with 72.46% F1-score.

The following figures show the performance of the models on all the examined datasets. Figure 2 shows how the models performed with a window size of 1 tempo, on each engine speed, Fig. 3 shows the results for windows size of 2 tempos and Fig. 4 for a window size of 5 tempos. All figures show the models' performance according to their F1-scores, during the validation phase of the experiments.

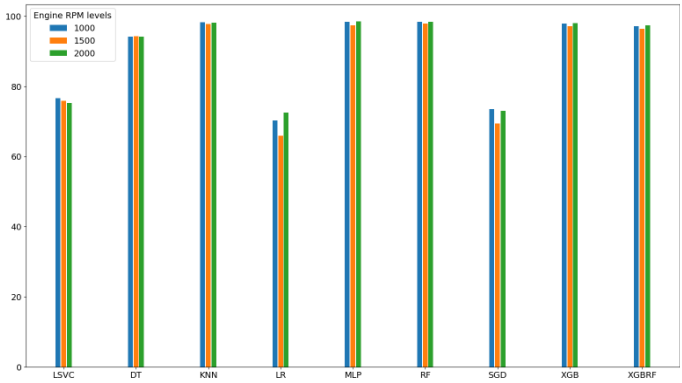

Fig. 2. F1-score of models for dataset with a window size of 1 tempo.

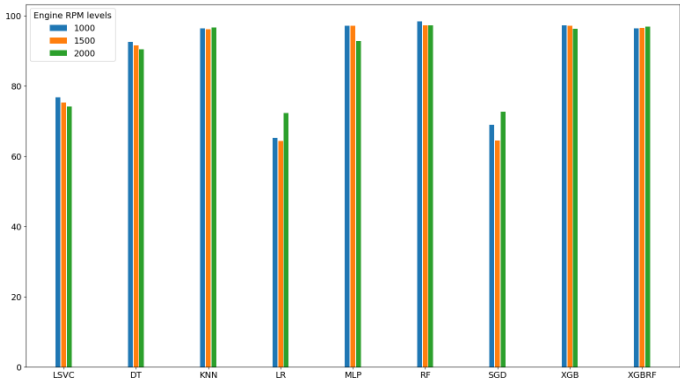

Fig. 3. F1-score of models for dataset with a window size of 2 tempo.

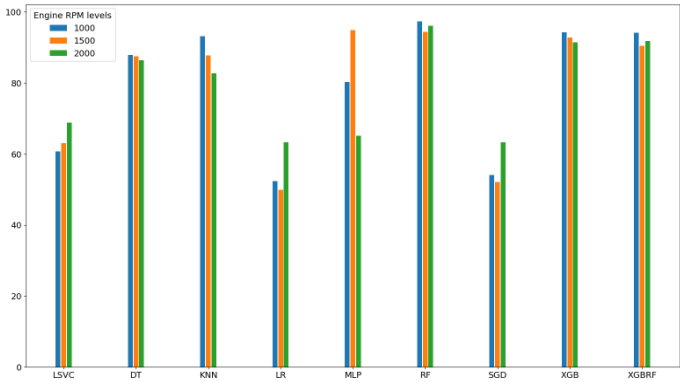

Fig. 4. F1 -score of models for dataset with a window size of 5 tempo.

From the figures, it is obvious that the higher the window size, the worse the models perform, with a small difference between window size of tempo 1 and 2 and a bigger difference when the window size is increased to 5 tempos. This is mostly due to the fact that the dataset becomes smaller when the window size increased, with the dataset having 1277 samples in the first case and only 277 in the second one. Moreover, the higher the rpm speed of the engine, the better the models perform, with some exceptions. Although the difference is small in most cases, the models perform better with the engines running at 2000 rpm compared to the other levels, with the worst performance being at 1500 rpm.

## V. Conclusions

In this study, we emphasized the future need for the identification of machines, in Smart City environments. The new paradigm of *machine biometrics*, which aims to identify machines in the context of smart city environments, as is done with humans until now. Moreover, the *engine biometric* was introduced based on the sound characteristics of a car engine, for predicting the car manufacturer. Additionally, a new dataset is presented to the literature, including 122 soundwaves of 19 different car models, belonging to 5 different car manufacturers, each one recorded in 3 different rpm levels. To our knowledge, it is the first dataset formed with specific acquisition protocol.

As for future work, the expansion of the dataset is required according to all aspects (manufacturers, models, and samples for each class). Moreover, as the application of deep learning has exponentially grown, their application should also be evaluated. Finally, an in-depth analysis of the permanence property of the proposed *engine biometric* will further improve its utility in practice.

ACKNOWLEDGMENT

This work was supported by the MPhil program "Advanced Technologies in Informatics and Computers", hosted by the Department of Computer Science, International Hellenic University, Greece.